\pdfoutput=1

\documentclass[11pt,a4paper]{article}
\usepackage{times,latexsym}
\usepackage{url}
\usepackage[T1]{fontenc}

\usepackage[acceptedWithA]{tacl2021v1}

\usepackage{xspace,mfirstuc,tabulary}

\newif\iftaclinstructions
\taclinstructionsfalse 
\iftaclinstructions
\renewcommand{\confidential}{}
\renewcommand{\anonsubtext}{(No author info supplied here, for consistency with
TACL-submission anonymization requirements)}
\newcommand{\instr}
\fi

\iftaclpubformat 

\else

\fi

\usepackage[utf8]{inputenc}

\usepackage{microtype}

\usepackage{inconsolata}

\usepackage{graphicx}
\usepackage{wrapfig}
\usepackage{amssymb}
\usepackage{color,xcolor,colortbl}
\usepackage{enumitem}
\usepackage{soul}
\usepackage{amsmath}
\usepackage{url}
\usepackage{algorithm}
\usepackage[noend]{algpseudocode}
\usepackage{booktabs}
\usepackage{bm,bbm}
\usepackage{mathtools}
\usepackage{array}
\usepackage{multirow}
\usepackage{subcaption}
\usepackage{pbox}
\usepackage{pifont}
\usepackage{arydshln}
\usepackage{dblfloatfix}
\usepackage{relsize}
\usepackage{setspace}
\usepackage[scaled=.8]{beramono}
\usepackage{listings}

\definecolor{darkblue}{rgb}{0.0, 0.0, 0.55}

\lstdefinestyle{myPDDLStyle}{
  basicstyle=\ttfamily\footnotesize,
  keywordstyle=\color{blue},
  identifierstyle=\color{black},
  commentstyle=\color{gray},
  stringstyle=\color{orange},
  showstringspaces=false,
  frame=single,
  numbers=left,
  numberstyle=\tiny\color{gray},
  breaklines=true,
  postbreak=\mbox{\textcolor{red}{$\hookrightarrow$}\space},
  tabsize=2,
  captionpos=b,
}

\lstdefinestyle{ALFWorldStyle}{   
    commentstyle=\color{green},
    keywordstyle=\color{blue},
    numberstyle=\tiny\color{gray},
    stringstyle=\color{purple},
    basicstyle=\ttfamily\footnotesize,
    breakatwhitespace=false,         
    breaklines=true,                 
    captionpos=b,                    
    keepspaces=true,                 
    numbersep=5pt,                  
    showspaces=false,                
    showstringspaces=false,
    showtabs=false,                  
    tabsize=2,
    xleftmargin=0pt,
}

\lstdefinelanguage{PDDL}{
  morekeywords={define,domain,:requirements,:predicates,:action,:parameters,:precondition,:effect,and,not},
  sensitive=false,
  morecomment=[l]{;},
  morestring=[b]"
}

\lstdefinestyle{HumanEval}{
    basicstyle=\ttfamily\small,
    breaklines=true,
    tabsize=2,
    frame=single,
}

\title{LASP: Surveying the State-of-the-Art in Large Language Model-Assisted AI Planning}

\author{Haoming Li,\Thanks{These authors contributed equally to this work.} \, Zhaoliang Chen,$^*$   Jonathan Zhang, Fei Liu \\
Department of Computer Science, Emory University\\
\texttt{\{haoming.li, david.chen2, jonathan.zhang2, fei.liu\}@emory.edu}}

\date{}

\begin{document}
\maketitle
\begin{abstract}
Effective planning is essential for the success of any task, from organizing a vacation to routing autonomous vehicles and developing corporate strategies. It involves setting goals, formulating plans, and allocating resources to achieve them. LLMs are particularly well-suited for automated planning due to their strong capabilities in commonsense reasoning. They can deduce a sequence of actions needed to achieve a goal from a given state and identify an effective course of action. However, it is frequently observed that plans generated through direct prompting often fail upon execution. Our survey aims to highlight the existing challenges in planning with language models, focusing on key areas such as embodied environments, optimal scheduling, competitive and cooperative games, task decomposition, reasoning, and planning. Through this study, we explore how LLMs transform AI planning and provide unique insights into the future of LM-assisted planning.
\end{abstract}

\section{Introduction}
\label{sec:intro}

Integrating LLMs into AI planning and decision-making systems is increasingly pertinent. However, many current methods are limited to constrained domains, such as household tasks performed by robots~\cite{shridhar2020alfworld,valmeekam2023planbench}, puzzles~\cite{yao2023treethoughtsdeliberateproblem}, maze navigation~\cite{silver2023generalized,lehnert2024abetterplanningtransformers}, and autonomous vehicle navigation~\cite{pan2024vlpvisionlanguageplanning}. The complex nature of real-world scenarios, such as trip planning and corporate strategic planning, calls for new frameworks that leverage LLMs' extensive world knowledge and commonsense reasoning. To address this gap, our survey critically examines existing planning benchmark datasets and methods. We seek to highlight promising directions and identify significant obstacles, offering a potential roadmap for harnessing LLMs' capabilities to tackle real-world planning challenges.

Planning involves generating a sequence of actions to achieve specific goals~\cite{russell95ai}. As illustrated in ALFWorld~\cite{shridhar2020alfworld}, a household robot performs a series of actions, e.g., `\emph{goto the stove}' and `\emph{take the pan from the stove}' to achieve the goal of `\emph{put a pan on the dining table.}' A key tool in this area is the Planning Domain Definition Language (PDDL; \citealt{Ghallab98}), which defines the necessary conditions before actions can take place and the effects of those actions afterwards. PDDL's symbolic representation of states and actions also resonates with formal semantics in computational linguistics~\cite{banarescu-etal-2013-abstract,ogorman-etal-2018-amr}. Our paper reviews key studies using PDDL in planning systems and discusses emerging alternatives that are suitable for handling open-domain tasks.

The quality and variety of benchmarks significantly influence the development of planning systems. We categorize existing benchmarks into three groups: (a) \textbf{\emph{embodied environments}}, where agents perform household tasks or navigate mazes, with algorithms designed to find the most efficient route through a grid~\cite{gupta2010blocks,shridhar2020alfworld,lehnert2024abetterplanningtransformers}; (b) \textbf{\emph{puzzle challenges}}, e.g., the Game of 24, graph coloring, and Towers of Hanoi, which increase in complexity as the problem size grows~\cite{valmeekam2023planbench,yao2023treethoughtsdeliberateproblem}; and (c) \textbf{\emph{natural language planning}} that focuses on optimal scheduling, travel planning, and task decomposition, which require advanced reasoning capabilities from LLMs~\cite{xie2024travelplannerbenchmarkrealworldplanning,zheng2024naturalplanbenchmarkingllms}. It should be noted that our survey does not include planning problems that require significant visual processing, such as autonomous driving~\cite{hu2023planningorientedautonomousdriving}, as these are beyond the scope of this study.

The ability to plan is crucial for embodied agents and robotic systems because these systems must break down high-level goals into a sequence of low-level, admissible actions to function effectively. Classical algorithms, such as Fast-Downward \cite{Helmert_2006}, can handle problems defined in PDDL, but many real-world applications are less ideal. Several issues arise, including goals that may be ambiguously expressed in natural language, environments that may only be partially observable, and state transition functions that may not be well-defined and are non-deterministic. LLMs offer a promising solution to some of these challenges, thanks to their flexibility and intrinsic understanding of the world.

Our survey addresses a crucial gap by examining the state-of-the-art planning methods and benchmarks. LLMs demonstrate \emph{text planning} capabilities, as they implicitly perform content determination and structuring during text generation~\cite{hovy-1988-planning,stent-etal-2004-trainable,moryossef-etal-2019-step}. Nevertheless, \emph{real-world planning} problems are versatile, spanning a range of contexts and may require an understanding of constraints. 

This survey is intended for NLP researchers interested in exploring AI planning problems. Junior researchers may find it beneficial to learn about these benchmarks and contribute new findings. Senior researchers will likely appreciate this survey, as it summarizes successful cases of integrating LLMs into planning frameworks, identifies opportunities for improvement, and encourages further contributions to the field of LLM-assisted planning (\textbf{\emph{LASP}}). The following sections provide a mathematical formulation of planning problems, discuss the Planning Domain Definition Language (PDDL), and survey popular benchmarks and methods.

\begin{table*}
\centering
\setlength{\tabcolsep}{5pt}
\renewcommand{\arraystretch}{1.1}
\begin{footnotesize}
\begin{tabular}{p{1.1in}p{4.9in}}
\toprule
\textbf{Blocksworld} \cite{gupta2010blocks} & Blocksworld, one of the earliest high-level task planning domains, was first introduced in SHRDLU. This domain typically involves multiple blocks on a table, with the objective being to reorganize them into a specified arrangement. A block must be at the top of a stack to be moved, and only one block can be moved at a time. Initially, Blocksworld supported four actions: `pick-up' and `put-down' for grasping and releasing a block from the table, and `stack' and `unstack' for moving blocks from and to other blocks. Despite various modifications introducing different constraints, the foundational actions of the original Blocksworld remain central to its operations.\\
\midrule
\textbf{VirtualHome} \cite{puig2018virtualhome} & VirtualHome is a collection of video simulations showcasing household tasks in eight different household scenes. Each entry in this dataset consists of a natural language description of an activity and its corresponding symbolic representation, called "programs," which outline the steps involved. The dataset contains two subsets of programs: those written by humans and those generated synthetically by the simulator. The human-written programs include 2,821 examples, with an average of 11.6 steps each. The synthetic programs, on the other hand, total 5,193 and average 9.6 steps per program. The content for each program is diversely generated, considering different homes, agents, camera perspectives, and the arrangement of some objects within the home.\\
\midrule
\textbf{ALFWorld} \cite{shridhar2021alfworldaligningtextembodied} & ALFWorld focuses on household tasks, using goals such as ``\emph{put a pan on the dining table}.'' It features a text-based environment (TextWorld) that uses PDDL semantics to produce textual observations and supports high-level text actions such as ``\emph{open a cabinet}'' or ``\emph{goto the stove}.'' Meanwhile, the embodied simulator (ALFRED) renders the world visually and interacts with low-level physical actions as from a robot, such as \texttt{MoveAhead} or \texttt{RotateLeft/Right}. Studies suggest that agents trained on text-based policies before executing in the embodied environment perform tasks faster and achieve better results than those trained directly in the embodied setting.\\
\midrule
\textbf{TEACh} \cite{padmakumar2022teach} & TEACh is a dataset comprising 3,215 dialogues that simulate a user interacting with a robot to perform household tasks. It features a Commander, equipped with oracle task details, who communicates with a Follower through natural language. The Follower executes the tasks and engages in dialogue to clarify instructions. Unlike planner-based simulations, these are human-human conversations that intertwine dialogue messages with actions taken in the environment. The aim of this benchmark is to enhance models' capabilities in language grounding, dialogue comprehension, and task execution by learning from human interactions.\\
\midrule
\textbf{ByteSized32-SP} \cite{wang2023bytesized32corpuschallengetask} & ByteSized32 is a corpus of 32 reasoning-focused text games totaling 20k lines of Python code. ByteSized32-SP expands on the original ByteSized32 by incorporating 76,369 state transitions to evaluate the LLM's performance in simulating state transitions. An LLM's task is to predict the next state based on the previous state provided in JSON format, the previous action, and a context message. The transitions are of two types: \emph{action}-driven and \emph{environment}-driven. An action-driven transition occurs, e.g., when turning on a sink, this updates the sink's property to ``\texttt{isOn=true}.'' An environment-driven transition occurs when the sink is turned on and water fills the cup, e.g., ``\texttt{\{"name":"cup", ..., "contains":["water"]\}}''. They show that LLMs are not reliable to accurately simulate environment-driven transitions and transitions that require arithmetic, commonsense, or scientific knowledge.\\
\midrule
\textbf{PlanBench} \cite{valmeekam2023planbench} & PlanBench introduces 8 test cases designed to evaluate LLMs' planning capabilities, covering aspects such as plan generation, cost-optimal planning, plan verification, execution reasoning, goal reformulation, plan reuse, replanning, and generalization. The initial dataset includes 600 instances from Blocksworld, i.e., stacking blocks on a table, and 285 from the Logistics domain, i.e., moving packages between locations using trucks or planes. In each case, LLMs are tested with few-shot prompting. PlanBench also uses obfuscated instances, where action names, predicate names, and object names are replaced with misleading terms to challenge the LLMs to plan without relying on commonsense knowledge. A standard planner would produce identical results for obfuscated instances.\\
\midrule
\textbf{Natural Plan} \cite{zheng2024naturalplanbenchmarkingllms} & Natural Plan is a benchmark consisting of three tasks: trip planning, meeting planning, and calendar scheduling. It evaluates LLMs' ability to handle planning tasks described in natural language. The data are collected from Google Flights, Google Maps, and Google Calendar, focusing on realistic scenarios. Natural Plan is a challenging benchmark, as evidenced by performances from leading models: GPT-4 and Gemini 1.5 Pro achieved solve rates of only 31.1\% and 34.8\% in trip planning, respectively. Model performance also tends to decrease as task complexity increases. \\
\bottomrule
\end{tabular}
\end{footnotesize}
\caption{A compilation of existing benchmarks developed for automated planning.}
\label{tab:benchmarks}
\end{table*}

\section{Classic Planning}
\label{sec:problem-formulation}

\paragraph{Problem Formulation.} In classic planning, an agent operates within a fully observable environment is modeled as a Markov Decision Process (MDP). The planning task involves a series of states and actions $(s_0, a_0, s_1, \ldots, a_{T-1}, s_T)$, beginning at an initial state $s_0$ and finally reach the goal state $s_T$. The environment is described by a set of states $\mathcal{S}$, and a set of actions $\mathcal{A}$ available to the agent. The state transition is modeled as $p_{\theta}(s_{t+1} | s_t, a_t)$, predicting the next state $s_{t+1}$ after an action $a_t$ is taken. This state transition modeling is sometimes known as a \emph{world model}, which is a representation of the environment that large language models can simulate. A reward function $\mathcal{S} \times \mathcal{A} \to \mathbb{R}$ assigns a scalar reward $r_\theta(r_t | s_t, a_t)$ after an action is taken from a given state. The ultimate goal of an MDP is to develop a policy, denoted as $a_t = p_{\phi}(a|s_t)$, focuses on identifying the optimal action $a_t$ given the current state $s_t$, that maximizes the total expected reward over time.

\paragraph{Domain Description.}

The Planning Domain Definition Language (PDDL; \citealt{Ghallab98}) is a representation used to define planning problems and domains using the BNF syntax. In PDDL, there are three main parts: a domain file, a problem file, and a plan. The \emph{domain} file describes the actions and predicates that can be used in various problems, the \emph{problem} file defines specific initial conditions and goals for a particular situation, and the \emph{plan} lists the actions needed to achieve the goals under those conditions. The following is an example of a PDDL for the Blocksworld domain. \emph{pick-up} and \emph{put-down} are two example actions. The preconditions and the post effects of these actions are specified with \emph{:prediction} and \emph{:effect}, which is a conjunction of predicates, indicating conditions can be true or false. 

PDDL offers the advantage of verifying whether plans produced by LLMs can actually be executed, as it can identify when preconditions, such as those in \emph{(unload b1 b2)}, are not met. However, its inflexibility and the need for creating domain and problem files pose significant drawbacks. Integrating PDDL with LLMs often involves converting these files and the plans into natural language, which can complicate the process.

\begin{lstlisting}[language=PDDL, style=myPDDLStyle, 
numbers=none, frame=lr, framesep=8pt, framerule=0pt]
(define (domain simplified_blocks)
  (:requirements :strips)
  (:predicates 
    (ontable ?x)
    (clear ?x)
    (handempty)
    (holding ?x))
    
  (:action pick-up
    :parameters (?x)
    :precondition (and (clear ?x) (ontable ?x) (handempty))
    :effect
      (and (not (ontable ?x))
           (not (clear ?x))
           (not (handempty))
           (holding ?x)))
           
  (:action put-down
    :parameters (?x)
    :precondition (holding ?x)
    :effect
      (and (not (holding ?x))
           (clear ?x)
           (handempty)
           (ontable ?x))))
\end{lstlisting}

\section{Planning Benchmarks}
\label{sec:benchmarks}

\subsection{Planning in Embodied Environments}

Embodied environments used to evaluate planning systems typically employ discrete action spaces and are narrowly limited to home tasks. 

\citet{shridhar2020alfredbenchmarkinterpretinggrounded} introduced \textbf{ALFRED} (Action Learning From Realistic Environments and Directives), a dataset that pairs natural language instructions with egocentric visual inputs to guide action sequences for household tasks. ALFRED comprises 25,743 English instructions covering 8,055 expert demonstrations, each with a high-level goal and an average of 50 steps, markedly expanding the scope of prior datasets. The dataset is split into training (21,023 annotations), validation (1,641 annotations), and testing (3,062 annotations) sections. The validation and testing segments are further divided into 'seen' and 'unseen' categories to evaluate how well models perform in new environments and with unfamiliar object types.

In evaluating models, ALFRED uses two metrics: Task Success and Goal-Condition Success. Task Success is binary, assigned a 1 if an object ends up in the right location and state to meet the task's requirements, and a 0 otherwise. Goal-Condition Success calculates the percentage of completed requirements at the end of a task relative to what was necessary for completion.

In \textbf{ALFWorld}~\cite{shridhar2020alfworld}, a model named BUTLER has been developed for evaluation on the dataset. This model comprises three components: brain, vision, and body. The brain element processes textual data, constructing a text-based conceptualization of the physical world using the TextWorld engine. Simultaneously, the vision component translates each visual scene from the physical environment into textual descriptions, employing a pre-trained Mask R-CNN detector at each step. The body component then translates high-level textual instructions into specific actions within the physical domain. In addition to BUTLER, a Seq2Seq model serves as the baseline for comparison.

The success rate is used to measure model performance across six task types, with BUTLER performing better than the baseline model. While BUTLER has perfect state estimation, object detection, and navigation, it does not utilize this advanced capability to improve its success rate and actually shows a decrease in performance. Nevertheless, when assessed with human-annotated goals, BUTLER achieves a notable success rate that may warrant further investigation. The best performance in this benchmark is achieved by ITCMA~\cite{zhang2024itcma}, which attains completion rates of 100\% on the seen set and 98\% on the unseen set. The experiment uses 3,553 training task instances spanning six categories, along with 140 in-distribution (seen set) and 34 out-of-distribution (unseen set) evaluation task instances.

The \textbf{VirtualHome} Activity Dataset~\cite{puig2018virtualhomesimulatinghouseholdactivities} is a collection of video simulations showcasing household tasks in eight different household scenes. Each entry in this dataset consists of a natural language description of an activity and its corresponding symbolic representation, called "programs," which outline the steps involved. The dataset contains two subsets of programs: those written by humans and those generated synthetically by the simulator. The human-written programs include 2,821 examples, with an average of 11.6 steps each. The synthetic programs, on the other hand, total 5,193 and average 9.6 steps per program. The content for each program is diversely generated, considering different homes, agents, camera perspectives, and the arrangement of some objects within the home.

To evaluate the program, the LCS between the predicted program and the ground truth is calculated. This metric allows for gaps in the matched steps as long as they appear in the correct order. Accuracy is assessed by dividing the length of the correct subsequence by the maximum length of the programs being compared. Notably, these metrics do not account for whether the predicted program can be executed, which is why the percentage of executable programs is also reported. The state-of-the-art model, ToolkenGPT~\cite{hao2024toolkengpt}, shows a 68\% success rate and 82\% executability, slightly exceeding the 66.66\% human-evaluated correctness reported in the 2018 dataset.

\subsection{Planning for Optimal Scheduling}
\label{sec:scheduling}

Planning is necessary for optimal scheduling, as it ensures that time and resources are properly managed, tools are used as needed, and intended goals are achieved within set constraints. Datasets have been developed to help with \emph{trip planning}, \emph{meeting scheduling}, \emph{calendar management}, \emph{auction bidding}, and \emph{logistics coordination}~\cite{valmeekam2023planbench,xie2024travelplannerbenchmarkrealworldplanning,chen2024moneymouthisevaluating,zheng2024naturalplanbenchmarkingllms}. These tasks typically involve constraints related to \emph{time}, \emph{budget}, and \emph{resource allocation}, e.g., adhering to specific time frames, assigning meeting spaces, transportation, and accommodations. Geographic factors, including distance and accessibility, are also important. 

\textbf{\emph{TravelPlanner}}~\cite{xie2024travelplannerbenchmarkrealworldplanning} focuses on developing language agents that can search and generate trip plans in response to user queries. These agents can navigate through approximately 4 million online entries using six specialized tools: \emph{CitySearch}, \emph{AttractionSearch}, \emph{FlightSearch}, \emph{DistanceMatrix}, \emph{RestaurantSearch}, and \emph{AccommodationSearch}. The benchmark also includes 1,225 user queries accompanied by human-annotated reference plans. An example user query is: ``\emph{I'm traveling from Seattle to California from November 6 to 10, 2023. I have a budget of \$6,000. For lodging, I prefer an entire room, and the accommodations must be pet-friendly.}'' LLM agents are designed to create travel plans that not only meet user needs but also adhere to commonsense constraints.

\subsection{Competitive and Cooperative Games}
\label{sec:multi-agents}

Collaborative and competitive games serve as testing grounds for evaluating LLMs' abilities in strategic planning, resource allocation, risk management, and multi-agent behaviors as they work toward specific goals. Examples of such games include \emph{Rock-Paper-Scissors}, \emph{Tower of Hanoi}, \emph{Minecraft}, \emph{Public Goods}, \emph{Guess 2/3 of the Average}, \emph{Auction}, \emph{Bargaining} and more~\cite{wu2024smartplaybenchmarkllmsintelligent,huang2024fardecisionmakingllmsevaluating,chen2024moneymouthisevaluating}. Games are commonly used as benchmarks because they have well-defined objectives and quantifiable outcomes.

\textbf{\emph{SmartPlay}}~\cite{wu2024smartplaybenchmarkllmsintelligent} is comprised of six games: \emph{Rock-Paper-Scissors}, \emph{Tower of Hanoi}, \emph{Two-armed Bandits}, \emph{Messenger}, \emph{Crafter}~\cite{hafner2022benchmarkingspectrumagentcapabilities}, and \emph{Minecraft}~\cite{fan2022minedojobuildingopenendedembodied}, all of which are accompanied by language descriptors. These games have been selected to challenge LLMs on essential capabilities such as reasoning with object dependencies, long-term planning, spatial reasoning, learning from history, and understanding randomness. 

\textsc{AucArena}~\cite{chen2024moneymouthisevaluating} is an evaluation suite designed to simulate multi-round auctions for a list of items with varying values. Items have a starting bid (e.g., \$1,000) and a true resale value (e.g., \$2,000). Each bidder operates with a budget (e.g., \$20,000) and tasked with maximizing their profits (their strategies can vary, such as acquiring specific items or securing as many items as possible.) In each round, agents bid on an item transparently, and the highest bidder in the final round wins the item. To succeed, bidder agents must operate with a fixed budget and make strategic, long-term decisions across several rounds.

\textbf{\emph{GAMA-Bench}}~\cite{huang2024fardecisionmakingllmsevaluating} features eight games divided into three categories: collaborative (\emph{Guess 2/3 of the Average}, \emph{El Farol Bar}, \emph{Divide the Dollar}), betrayal (\emph{Public Goods Game}, \emph{Diner's Dilemma}, \emph{Sealed-Bid Auction}), and sequential (\emph{Battle Royale}, \emph{Pirate Game}). These games have been studied in Game Theory literature. For example, in the \emph{Public Goods Game}, each player receives an amount of money and can choose how much to contribute to a common pot. The total amount in the pot is then multiplied (usually doubled) and distributed equally among all participants, regardless of individual contribution. \emph{The Nash equilibrium in this game is for all players to contribute nothing, as each player hopes to free-ride on the contributions of others.} An LLM agent is then evaluated by how its contributions nearly sum to zero while interacting with others.

\subsection{Task Decomposition}
\label{sec:task-decompose}

Task decomposition benefits planning by enabling efficient and reliable execution. Breaking down a task into subtasks facilitates the creation of a task-specific taxonomy. As a result, tasks can often be executed more effectively when provided with a concrete plan that includes actionable steps. It also becomes easier to recover from interruptions~\cite{yuan2023tasklamaprobingcomplextask,yuan2023distillingscriptknowledgelarge,ou2024worldapisworldworthapis}.

\textbf{\emph{TaskLAMA}}~\cite{yuan2023tasklamaprobingcomplextask} has developed a dataset comprising 1,612 annotated complex tasks, which includes 711 tasks from the MSComplexTasks dataset~\cite{zhang2021learning} and 901 tasks derived from how-to search queries. Examples of these tasks are ``\emph{cook lobster tails at home (Grilled)}'' and ``\emph{plan a wedding (In Italy)}.'' Human annotators are tasked with: 1) writing their assumptions to contextualize the task (denoted in parentheses); 2) outlining the necessary steps for completing the tasks within this context; 3) detailing the temporal dependencies among these steps. The resulting task is structured into a directed acyclic graph known as a Task Graph, where each node represents a step, and the edges indicate the temporal dependencies between these steps.

\textsc{WorldAPIs}~\cite{ou2024worldapisworldworthapis} uses a top-down strategy to derive APIs (actions) from wikiHow's step-by-step instructions for everyday tasks. For example, the task ``\emph{How to Melt Chocolate in Microwave}'' can be broken down into steps such as ``\emph{Chop the chocolate}'' and ``\emph{Place the chocolate},'' ending with ``\emph{Allow the chocolate to cool}.'' Starting with an initial set of APIs, WORLDAPIS employs the LLM to iteratively generate Python programs for these tasks. When existing APIs cannot cover a step, the program `hallucinates' new APIs, which are then added to the pool. This method has expanded the action space to over 300 APIs necessary for tasks in the physical world. In contrast, existing simulators support only a fraction (9 of the top 50) of these induced APIs.

\citet{yuan2023distillingscriptknowledgelarge} introduce the task of \emph{constrained language planning} and present \texttt{CoScript}, a dataset comprising 55,000 goal-oriented scripts. Each script is a sequence of the necessary steps to achieve a specific goal. For instance, to achieve the goal of `\emph{make a cake,}' one may follow steps such as \emph{gathering ingredients} and \emph{preheating the oven}. Additionally, constrained language planning imposes various constraints on planning goals. E.g., a cake can be made using different ingredients (e.g., chocolate or vanilla), various tools (a microwave or an oven), or for distinct purposes (a wedding or a birthday party). The authors use an ``over-generate-then-filter'' approach to select high-quality scripts from multiple LLM-generated candidates. A good planner is expected to generate steps that respect these constraints. 

\subsection{Reasoning and Planning}

Reasoning and planning are distinct in terms of their focus. Reasoning involves integrating multiple pieces of information and making inferences to address complex problems. E.g., `\emph{What musical instruments do Minnesota-born Nobel Prize winners play?}' is considered a multi-hop reasoning problem as it involves inferential chaining to generate an answer. In contrast, planing is not only about achieving goals but doing so optimally, e.g., with minimal cost or shortest path. Planning involves considering various constraints and predicting future states that result from actions. E.g., `\emph{booking the cheapest flight}' involves generating a sequence of actions (search, compare, book) around a clear objective with temporal and cost constraints. 

Despite the distinction, reasoning can still play a secondary role in planning. For example,
\textsc{PrOntoQA}~\cite{saparov2023languagemodelsgreedyreasoners} is a synthetic QA dataset designed to assess LLMs' reasoning capability. This dataset is based on synthetic world models represented in first-order logic. Each sentence in the chain-of-thought is parsed into a formal representation to reconstruct proof steps, which are evaluated against a gold-standard proof. Their study suggests that while LLMs can effectively perform individual deduction steps, they struggle with planning. Particularly, LLMs face challenges in selecting the correct proof step where multiple viable options are available, often leading to incomplete proofs and incorrect answers.

\textsc{AgentBench}~\cite{liu2023agentbenchevaluatingllmsagents} evaluates LLMs' reasoning and decision-making in a multi-turn, open-ended context. It is designed for text-only LLMs acting as autonomous agents, and features 8 distinct environments, including \emph{operating system}, \emph{database}, \emph{knowledge graph}, \emph{digital card game}, \emph{lateral thinking puzzles}, \emph{housekeeping}, and \emph{web shopping and browsing}. The benchmark focuses on LLMs' core skills such as instruction following, knowledge acquisition, logical reasoning, and commonsense grounding. Results suggest that training LLMs on code and high-quality, multi-turn alignment data enhances agent performance.

\textsc{SWE-bench}~\cite{jimenez2024swebenchlanguagemodelsresolve} is an evaluation framework that includes 2,294 software engineering problems collected from 12 Python GitHub repositories. It presents a codebase and issue description for an LLM to solve. Solutions are then evaluated using the repository's existing testing framework. Resolving issues in \textsc{SWE-bench} requires LLMs to modify various functions, classes, or files, challenging their ability for processing long contexts and complex reasoning.

\section{LLM-Assisted Planning Methods}
\label{sec:methods}

We provide a modularized view of LLM-assisted planning algorithms, which includes the \textit{plan generator}, \textit{environment interpreter}, and \textit{enhanced feedback provider}. 
Our goal is not to survey all of the planning methods. Instead, we focus on undersanding LLMs' role in assisting the development of SOTA planning systems and to solve new planning problems. We broadly categorize the methods into LLM-as-Planner and LLM-as-Facilitator. The first category explicitly uses LLM's inherit reasoning abilities to generate plans, while the latter relies on other planning algorithms for plan generation and LLMs only serve to facilitate the process, such as as a simulator for the world model, or action planner to estimate the future actions from the current state, or using LLMs to redefine the action space with world APIs. Note that we provide an in-depth discussion on whether LLMs can plan as its intrinsic ability in the discussion section. 

\textit{Plan generator} is the core of the entire operation, which predicts one future action, multiple future actions, or even multiple threads of future actions. Many recent works used LLMs as the plan generator, exploiting its versatility to handle scenarios and problems across different domains. However, some researchers argued that LLMs in their current states are fundamentally weak at planning. So, many have also proposed using algorithms such as Fast-Downward \cite{Helmert_2006} and even separately trained models to perform planning \cite{liu2024learning}. We argue that the exploration of plan generators, especially non-LLM based planners, are far from an end. Recently, for example, \cite{lehnert2024abetterplanningtransformers} proposed Searchformer, a transformer based planning algorithm that was both performant and incredibly efficient.

\subsection{Using LLM as an Action Planner}
\label{sec:llm-planner}

Using LLM as planner relies heavily on the prompt design and LLM’s inherent ability to generate and refine plans. The key advantage of using LLMs to generate plans lies in their ability to understand and generate natural language, allowing them to process problem descriptions and generate plans without the need for extensive domain-specific knowledge or specialized training. One main disadvantage, however, is that LLMs are inherently non-deterministic and its behavior is much harder to predict compared to using tools such as symbolic planners. 

\subsubsection{Dynamic Plan Updates with Feedback}

Reflexion \cite{shinn2023reflexion} employs an iterative reinforcement process where agents generate actions, receive evaluations, and produce reflective feedback to improve subsequent actions. In particular, this system features a memory module where short-term memory contains the recent actions and their outcomes, while long-term memory stores linguistic reflections which guide the future behavior. Compared to sparse reward signals, linguistic self-reflection provides richer information and context to help agent to navigate complex planning problems. 
ReAct \cite{yao2023react} integrates reasoning and acting in LLMs to enhance decision-making in language and interactive tasks. By interleaving reasoning traces and actions, the system can dynamically update plans and interact with external sources to refine its actions, which helps mitigate issues like hallucination and error propagation seen in traditional models. 

\citet{huang2022language} proposes a method called {Translated ${\langle LM\rangle}$}, which employs a multi-step planning process where large language models first generate free-form action plans, which are then semantically translated into admissible actions using a pre-trained BERT-style model. The plans are generated autoregressively, with each step corrected and conditioned on previously translated actions. Dynamic example selection further refines the initial prompts, ensuring relevant contextual examples guide the model, similar to in-context learning setup.

LLM-Planner \cite{song2023llmplanner} employs large language models to perform few-shot grounded high-level planning for embodied agents. It generates high-level plans from natural language instructions and dynamically re-plans based on environmental observations.
G-PlanET \cite{lin2023gplanet} employs a method where LMs take a high-level task description and an object table from a realistic environment as inputs, generating a step-by-step plan using an iterative decoding strategy. This approach involves flattening object tables into token sequences and integrating them with seq2seq learning frameworks. This approach does not use a decoder-only language model such as GPT, but rather an encoder-decoder model known as BART. 

Inner Monologue \cite{huang2022inner} uses a continuous feedback loop where an LLM processes various types of textual feedback from the environment -- including success detecftion, passive scene description, and active scene description -- to enhance planning and reasoning in robotic tasks. This method allows for real-time adjustments and improvements in action sequences based on success detection, scene descriptions, and human interactions.
ISR-LLM \cite{zhou2023isrllm} improves LLM-based long-horizon planning by using an iterative self refinement process. Initially, natural language instructions are translated into PDDL files, which are then used to generate an action plan. This plan is validated and refined iteratively based on feedback. The PDDL translation and plan generation are both performed by LLMs, and validation phase is done with either an LLM or an external mechanism. 

Self-Refine \cite{madaan2023selfrefine} employs an iterative process where an LLM generates an initial output, provides self-feedback, and refines the output based on this feedback. Here, LLM serves as plan generator, refiner, and feedback provider. The feedback provider is prompted to generate actionable and specific feedback. 
\citet{zhao2023explicit} argues that during inferring, planning will be able to find more useful facts which could lead to success reasoning. Planning-based reasoning systems are easier to interpret and therefore tend to be more useful in user-centred and safety-critical scenarios. 

SELFGOAL~\cite{yang2024selfgoallanguageagentsknow} aims to break down high-level, non-executable goals such as ``\emph{maximizing profit}'' actionable subgoals. Their method was tested in four games requiring multiple agents, including the \emph{Public Goods},\emph{Guess 2/3 of the Average}, \emph{First-price Auction}, and \emph{Bargaining}. In these games, agents may deviate from their goals without tangible subgoals or if these subgoals are not grounded in the environment.

It has three main modules: Search, Decompose, and Act, all powered by a LLM. In the Decompose stage, the system uses the LLM's prior knowledge to break down a high-level goal into a tree of practical subgoals. It then selects the K most effective subgoals that contribute towards achieving the main objective. In the Act stage, these chosen subgoals guide the LLM in generating specific actions based on the current scenario. This method seeks to balance \emph{prior task decomposition} and \emph{post-hoc experience summarization} to effectively realize high-level objectives. 

The Tree of Thoughts (ToT) \cite{yao2023treethoughtsdeliberateproblem} framework enhances LLMs' problem-solving capabilities by structuring the task as a tree search, where each node represents a partial solution or thought. This method enables the LLM to explore multiple reasoning paths, evaluate them heuristically, and apply systematic search algorithms like BFS or DFS. 
Reasoning via Planning (RAP) \cite{hao2023reasoning} employs a strategy where the LLM is repurposed as both a world model and a reasoning agent, incorporating Monte Carlo Tree Search (MCTS) to guide strategic exploration. The process involves incrementally building a reasoning tree under the guidance of the world model and rewards, effectively balancing exploration and exploitation to find high-reward reasoning paths. This process maintains a grounded and coherent inference by simulating future states, making it more reliable for complex tasks. 

SayCanPay \cite{hazra2024saycanpay} integrates LLMs' generative capabilities with heuristic search to produce feasible and cost-effective plans. The method involves three steps: generating candidate actions (Say), evaluating their feasibility (Can), and estimating their payoff (Pay). The approach employs both Greedy-Action and Beam-Action decoding strategies to optimize the planning process. SayCanPay's main advantage is its ability to combine LLM-generated actions with heuristic evaluation, leading to more grounded and efficient plans. However, the method requires extensive domain-specific training data and may struggle with out-of-distribution generalization.


\subsubsection{Code-based Planning and Prompting} 

PROGPROMPT \cite{singh2022progprompt} leverages a programming language-inspired prompt structure to guide LLMs in generating robot task plans that are executable and contextually appropriate. By incorporating assert statements and recovery actions, the method ensures that the plans are grounded in the current state of the environment and robot capabilities.

\citet{raman2022planning} also proposes a prompting-based method that leverages precondition errors to improve the generation of executable and semantically correct plans from LLMs. In particular, upon detecting a precondition error, the method re-prompts the LLM with information about the error to extract a corrective action. The types of information provided includes the notion of error, inference of error, and cause of error. However, they may not be always available, depending on the environment. 

\citet{silver2023generalized} proposed using LLMs as generalized planners. In this framework, GPT-4 is prompted to first summarize the PDDL domain, propose a non-search-based strategy, and then implement this strategy in Python which returns a list of actions. Automated debugging iteratively improves the code based on various types of feedback, including Python exceptions, timeouts, plan syntax, and plan semantics. 

AdaPlanner \cite{sun2023adaplanner} employs an adaptive closed-loop approach where an LLM agent generates and refines plans in response to environmental feedback, using Pythonic code prompts to minimize ambiguity. This framework performs refinement both using failure and success experiences. For in-plan feedback, where the environment is in line with the predicted plan, AdaPlanner extracts useful information from this success experience for upcoming actions. For Out-of-plan Feedback where the observation deciates from the predicted feedback, AdaPlanner refines the plan based on the previous checkpoint. 


\subsubsection{Using Separately Trained Components} 

The LID framework~\cite{li2022lid} uses pre-trained LMs to initialize a policy network, which receives goals, observations, and histories as inputs and outputs the prediction of the next action. The policy network was trained through imitation learning on expert demonstrations. However, expert data is not always available. To address this limitation, they also introduced Active Data Gathering (ADG), where the model explores the environment and learns from experience.  

DEPS~\cite{wang2023describe} employs a cyclic process of describing the current state, explaining errors, re-planning, and selecting the most feasible sub-goals. Planner and explainer are both LLMs, describer is a vision-language model (VLM) to handle multi-modality inputs, and selector is a separately trained module. This method allows for iterative improvement of plans, addressing the inefficiencies of static planning approaches. 

\subsubsection{Planning with Constraints}

In a planning problem, the implicit constraints include preconditions for a given action, the range of possible actions, and etc. However, sometimes constraints are explicit in the goal. \citet{yuan2023distillingscriptknowledgelarge} introduced a method targeted to generate scripts that adhere to specific constraints (such as making a cake for people with diabetes). The paper proposed an ``over-generate-then-filter approach'', which initially generates multiple script options and then filtering these to select the most suitable ones. This approach not only enhances the relevance of the scripts to the specified goals but also significantly improves the faithfulness to the constraints, a noted improvement over traditional models that often fail to consider finer details in constraints.

 
\subsection{LLM-as-Facilitator}
While using LLMs to generate plans offers remarkable flexibility, many have noted that LLMs often fail to produce feasible and optimal plans for complex, multi-step problems that involve understanding and manipulating the state of the world \cite{liu2023llmp}. Therefore, some only use LLMs as a facilitator to power other algorithms to generate the plan. 

\subsubsection{Using Symbolic Planners} 

LLM+P~\cite{liu2023llmp}, for example, employs a 3 step process. The LLM first takes a natural language description of a planning problem and translates it into a PDDL file, which is fed into a classical planner such as Fast-Downward \cite{Helmert_2006}. After the problem is solved using the planner, LLM translates the solution back in natural language for interpretability. One important assumption needed to use this system is that the domain PDDL file is readily available. In many real world problems, however, it is sometimes infeasible to always have a well-defined domain PDDL file, making this approach less flexible compared to methods introduced in LLM-as-Planner. 

LLM Dynamic Planner (LLM-DP)~\cite{dagan2023dynamic} tries to tackle this limitation.  employs a hybrid approach where an LLM interprets natural language instructions and generates PDDL goals, while a symbolic planner generates and executes plans to achieve these goals. Specifically, the process starts with the LLM converting natural language task instructions into a PDDL goal using fixed in-context examples. The initial scene is parsed to establish the initial World State (W) and Beliefs (B). The planning loop repeats until the goal is achieved. In this loop, the LLM samples beliefs to generate plausible world states, which are then used by the Plan Generator to create plans for each sampled world state. The Action Selector chooses the shortest valid plan from these. After executing an action, the system observes the environment and updates W and B based on the results, which may trigger re-planning if new information is discovered.

\citet{wong2023learning} proposed Action Domain Acquisition (Ada), which employs an iterative learning process to develop a library of high-level action abstractions and low-level controllers by leveraging natural language descriptions and interactive planning. It integrates LLMs for proposing symbolic action abstractions and goal definitions, and uses a hierarchical planner for grounding and executing plans. The high level planning is done using symbolic planner Fast-Downward \cite{Helmert_2006}.

Planning Abstraction from Language (PARL) \cite{liu2024learning} employs a three-stage process of discovering symbolic action spaces from language, learning  planning-compatible model, and utilizing these models in combination with a simple tree search (BFS) for planning and execution at inference time. Note that this approach does not rely on external symbolic planners. Rather, in the second stage it trains a set of 4 separate models: 1) state abstraction function that maps raw state to abstract state, 2) abstract transition function that models the transition in the abstract state space, 3) a feasibility function for each abstract action, and 4) low-level policy function that maps abstract actions to specific actions in raw action space. 

\citet{silver2022pddl} proposed LLM Plan Guidance method, which initializes the priority queue of a heuristic search planner, specifically a Greedy Best-First Search (GBFS) planner, using the output of an LLM. This initialization aims to provide a substantial head start for the planner by potentially skipping over difficult-to-navigate early states and moving quickly towards solution states.

\section{Discussion}

In this section, we explore some challenges of using LLMs for planning, including their inherent limitations, changes in behavior due to model updates, strategies for improving robustness, understanding physical constraints, and addressing latency issues.

\subsection{Can LLMs Plan?}

This topic has always been being discussed, and yet the conclusion is not determined. \citet{Kambhampati_2024} claims that as n-gram models, LLMs excel at universal approximate retrieval, an ability that is often confused with the capability to reason or plan. The results of an experiment, in which the names of objects in a planning problem were obscured, demonstrate a reduction in the models' ability to retrieve approximations, highlighting their lack of competence in planning, even in advanced models like GPT-4. Although the capability of LLMs in performing planning tasks remains controversial, they still play a constructive role. Due to their vast capacity for knowledge retention, the plans they generate can often be utilized cooperatively with an external verifier or within a human-in-the-loop process.

\citet{valmeekam2023planbench} focuses on the executability of plans generated by LLMs. They questioned whether LLMs can generate a valid or optimal plan and elucidate the inner logic of the plan by describing the conditions of each task. The ability to reuse previous plans has also been tested. Several experiments have been conducted using the PlanBench dataset and the results are somewhat disheartening. They observed that for the most advanced models, even minor changes in objects can cause significant decreases in performance. Additionally, although LLMs can generate a plan, they cannot explain what the plan actually accomplishes, suggesting that the model does not fully understand the task or the plan itself.

\citet{pallagani2023understanding} evaluate various LLMs across six classic planning domains, addressing four critical questions: (1) How effective are LLMs in generating plans? (2) Which \emph{pre-training data} best supports plan generation? (3) Can \emph{fine-tuning} and \emph{prompting} enhance LLMs' planning capabilities? (4) Are LLMs capable of \emph{generalizing} plans? They show favorable results of LLMs for automated planning with appropriate selection of the LLM, data preparation, and fine-tuning. However, in terms of \emph{generalization}, LLMs exhibit limited capabilities, whether for planning in unknown domains, handling randomly named objects, or generalizing across different plan lengths.

\subsection{Robustness of LLM-Assisted Planning}

LLM planners based on GPT-4 and other large-scale enterprise models can be unreliable due to potential behavior changes (snapshot updates) \& expensive pricing. Recent research suggests that groups of smaller language models, such as those with 6 billion and 13 billion parameters, can perform similarly to much larger counterparts like the GPT-3.5, which has roughly 175 billion parameters. According to \citet{lu2024blending}, using responses randomly selected from these smaller models—each conditioned on previous outputs from its counterparts—allows these systems to work together effectively. This collaboration can enhance user engagement. This approach could offer a more stable and cost-effective alternative to deploying single, large-scale models, which may change unpredictably with updates and be costly.

\citet{jiao2024learning} Jiao et al. (2024) address the latency issues associated with reasoning-as-planning (RAP) methods, which stem from the need to frequently evaluate intermediate reasoning states and navigate a large exploration space. They aim to mitigate these delays by reconceptualizing RAP as a learning problem and employing a Monte Carlo Tree Search (MCTS)-like method to simulate the value of each intermediate step, typically the most time-consuming part of RAP. While this approach reduces some latency, it still requires significant resources, indicating that further innovation in simulation strategies remains an open area for future research.

A remaining issue is LLMs' handling of physical constraints~\cite{curtis2024trustproc3ssolvinglonghorizon}, which sometimes leads to the generation of plans that cannot be executed. These systems can struggle with real-world dynamics and spatial relations, resulting in theoretical plans that don't work in practice. This gap can make it difficult for automatic planning to be applied in fields such as robotics and logistics, where accurate, context-sensitive planning is crucial.

\section{Conclusion}

In this paper, we focus on exploring various techniques of LLM Planning and review commonly used benchmarks. We present benchmarks that assess model performance across diverse settings and complexity levels. We discuss whether LLMs truly facilitate planning or merely retrieve data, and we examine potential robustness issues in LLM-assisted planning.

\bibliography{custom}

\begin{thebibliography}{65}
\expandafter\ifx\csname natexlab\endcsname\relax\def\natexlab#1{#1}\fi

\bibitem[{Banarescu et~al.(2013)Banarescu, Bonial, Cai, Georgescu, Griffitt, Hermjakob, Knight, Koehn, Palmer, and Schneider}]{banarescu-etal-2013-abstract}
Laura Banarescu, Claire Bonial, Shu Cai, Madalina Georgescu, Kira Griffitt, Ulf Hermjakob, Kevin Knight, Philipp Koehn, Martha Palmer, and Nathan Schneider. 2013.
\newblock \href {https://aclanthology.org/W13-2322} {{A}bstract {M}eaning {R}epresentation for sembanking}.
\newblock In \emph{Proceedings of the 7th Linguistic Annotation Workshop and Interoperability with Discourse}, pages 178--186, Sofia, Bulgaria. Association for Computational Linguistics.

\bibitem[{Chen et~al.(2024)Chen, Yuan, Ye, Majumder, and Richardson}]{chen2024moneymouthisevaluating}
Jiangjie Chen, Siyu Yuan, Rong Ye, Bodhisattwa~Prasad Majumder, and Kyle Richardson. 2024.
\newblock \href {http://arxiv.org/abs/2310.05746} {Put your money where your mouth is: Evaluating strategic planning and execution of llm agents in an auction arena}.

\bibitem[{Curtis et~al.(2024)Curtis, Kumar, Cao, Lozano-Pérez, and Kaelbling}]{curtis2024trustproc3ssolvinglonghorizon}
Aidan Curtis, Nishanth Kumar, Jing Cao, Tomás Lozano-Pérez, and Leslie~Pack Kaelbling. 2024.
\newblock \href {http://arxiv.org/abs/2406.05572} {Trust the proc3s: Solving long-horizon robotics problems with llms and constraint satisfaction}.

\bibitem[{Dagan et~al.(2023)Dagan, Keller, and Lascarides}]{dagan2023dynamic}
Gautier Dagan, Frank Keller, and Alex Lascarides. 2023.
\newblock Dynamic planning with a llm.
\newblock \emph{arXiv preprint arXiv:2308.06391}.

\bibitem[{Fan et~al.(2022)Fan, Wang, Jiang, Mandlekar, Yang, Zhu, Tang, Huang, Zhu, and Anandkumar}]{fan2022minedojobuildingopenendedembodied}
Linxi Fan, Guanzhi Wang, Yunfan Jiang, Ajay Mandlekar, Yuncong Yang, Haoyi Zhu, Andrew Tang, De-An Huang, Yuke Zhu, and Anima Anandkumar. 2022.
\newblock \href {http://arxiv.org/abs/2206.08853} {Minedojo: Building open-ended embodied agents with internet-scale knowledge}.

\bibitem[{Ghallab et~al.(1998)Ghallab, Howe, Knoblock, Mcdermott, Ram, Veloso, Weld, and Wilkins}]{Ghallab98}
M.~Ghallab, A.~Howe, C.~Knoblock, D.~Mcdermott, A.~Ram, M.~Veloso, D.~Weld, and D.~Wilkins. 1998.
\newblock \href {http://citeseerx.ist.psu.edu/viewdoc/summary?doi=10.1.1.37.212} {{PDDL---The Planning Domain Definition Language}}.

\bibitem[{Gupta et~al.(2010)Gupta, Efros, and Hebert}]{gupta2010blocks}
Abhinav Gupta, Alexei~A Efros, and Martial Hebert. 2010.
\newblock Blocks world revisited: Image understanding using qualitative geometry and mechanics.
\newblock In \emph{Computer Vision--ECCV 2010: 11th European Conference on Computer Vision, Heraklion, Crete, Greece, September 5-11, 2010, Proceedings, Part IV 11}, pages 482--496. Springer.

\bibitem[{Hafner(2022)}]{hafner2022benchmarkingspectrumagentcapabilities}
Danijar Hafner. 2022.
\newblock \href {http://arxiv.org/abs/2109.06780} {Benchmarking the spectrum of agent capabilities}.

\bibitem[{Hao et~al.(2023)Hao, Gu, Ma, Hong, Wang, Wang, and Hu}]{hao2023reasoning}
Shibo Hao, Yi~Gu, Haodi Ma, Joshua~Jiahua Hong, Zhen Wang, Daisy~Zhe Wang, and Zhiting Hu. 2023.
\newblock Reasoning with language model is planning with world model.
\newblock \emph{arXiv preprint arXiv:2305.14992}.

\bibitem[{Hao et~al.(2024)Hao, Liu, Wang, and Hu}]{hao2024toolkengpt}
Shibo Hao, Tianyang Liu, Zhen Wang, and Zhiting Hu. 2024.
\newblock \href {http://arxiv.org/abs/2305.11554} {Toolkengpt: Augmenting frozen language models with massive tools via tool embeddings}.

\bibitem[{Hazra et~al.(2024)Hazra, Zuidberg Dos~Martires, and De~Raedt}]{hazra2024saycanpay}
Rishi Hazra, Pedro Zuidberg Dos~Martires, and Luc De~Raedt. 2024.
\newblock Saycanpay: Heuristic planning with large language models using learnable domain knowledge.
\newblock \emph{arXiv preprint arXiv:2308.12682}.

\bibitem[{Helmert(2006)}]{Helmert_2006}
M.~Helmert. 2006.
\newblock \href {https://doi.org/10.1613/jair.1705} {The fast downward planning system}.
\newblock \emph{Journal of Artificial Intelligence Research}, 26:191–246.

\bibitem[{Hovy(1988)}]{hovy-1988-planning}
Eduard~H. Hovy. 1988.
\newblock \href {https://doi.org/10.3115/982023.982043} {Planning coherent multisentential text}.
\newblock In \emph{26th Annual Meeting of the Association for Computational Linguistics}, pages 163--169, Buffalo, New York, USA. Association for Computational Linguistics.

\bibitem[{Hu et~al.(2023)Hu, Yang, Chen, Li, Sima, Zhu, Chai, Du, Lin, Wang, Lu, Jia, Liu, Dai, Qiao, and Li}]{hu2023planningorientedautonomousdriving}
Yihan Hu, Jiazhi Yang, Li~Chen, Keyu Li, Chonghao Sima, Xizhou Zhu, Siqi Chai, Senyao Du, Tianwei Lin, Wenhai Wang, Lewei Lu, Xiaosong Jia, Qiang Liu, Jifeng Dai, Yu~Qiao, and Hongyang Li. 2023.
\newblock \href {http://arxiv.org/abs/2212.10156} {Planning-oriented autonomous driving}.

\bibitem[{Huang et~al.(2024)Huang, Li, Lam, Liang, Wang, Yuan, Jiao, Wang, Tu, and Lyu}]{huang2024fardecisionmakingllmsevaluating}
{Jen-tse} Huang, Eric~John Li, Man~Ho Lam, Tian Liang, Wenxuan Wang, Youliang Yuan, Wenxiang Jiao, Xing Wang, Zhaopeng Tu, and Michael~R. Lyu. 2024.
\newblock \href {http://arxiv.org/abs/2403.11807} {How far are we on the decision-making of llms? evaluating llms' gaming ability in multi-agent environments}.

\bibitem[{Huang et~al.(2022{\natexlab{a}})Huang, Abbeel, Pathak, and Mordatch}]{huang2022language}
Wenlong Huang, Pieter Abbeel, Deepak Pathak, and Igor Mordatch. 2022{\natexlab{a}}.
\newblock Language models as zero-shot planners: Extracting actionable knowledge for embodied agents.
\newblock In \emph{International Conference on Machine Learning}, pages 9118--9147. PMLR.

\bibitem[{Huang et~al.(2022{\natexlab{b}})Huang, Xia, Xiao, Chan, Liang, Florence, Zeng, Tompson, Mordatch, Chebotar et~al.}]{huang2022inner}
Wenlong Huang, Fei Xia, Ted Xiao, Harris Chan, Jacky Liang, Pete Florence, Andy Zeng, Jonathan Tompson, Igor Mordatch, Yevgen Chebotar, et~al. 2022{\natexlab{b}}.
\newblock Inner monologue: Embodied reasoning through planning with language models.
\newblock \emph{arXiv preprint arXiv:2207.05608}.

\bibitem[{Jiao et~al.(2024)Jiao, Qin, Liu, Chen, and Joty}]{jiao2024learning}
Fangkai Jiao, Chengwei Qin, Zhengyuan Liu, Nancy~F Chen, and Shafiq Joty. 2024.
\newblock Learning planning-based reasoning by trajectories collection and process reward synthesizing.
\newblock \emph{arXiv preprint arXiv:2402.00658}.

\bibitem[{Jimenez et~al.(2024)Jimenez, Yang, Wettig, Yao, Pei, Press, and Narasimhan}]{jimenez2024swebenchlanguagemodelsresolve}
Carlos~E. Jimenez, John Yang, Alexander Wettig, Shunyu Yao, Kexin Pei, Ofir Press, and Karthik Narasimhan. 2024.
\newblock \href {http://arxiv.org/abs/2310.06770} {Swe-bench: Can language models resolve real-world github issues?}

\bibitem[{Kambhampati(2024)}]{Kambhampati_2024}
Subbarao Kambhampati. 2024.
\newblock \href {https://doi.org/10.1111/nyas.15125} {Can large language models reason and plan?}
\newblock \emph{Annals of the New York Academy of Sciences}, 1534(1):15–18.

\bibitem[{Lehnert et~al.(2024)Lehnert, Sukhbaatar, Su, Zheng, Mcvay, Rabbat, and Tian}]{lehnert2024abetterplanningtransformers}
Lucas Lehnert, Sainbayar Sukhbaatar, DiJia Su, Qinqing Zheng, Paul Mcvay, Michael Rabbat, and Yuandong Tian. 2024.
\newblock \href {http://arxiv.org/abs/2402.14083} {Beyond a*: Better planning with transformers via search dynamics bootstrapping}.

\bibitem[{Li et~al.(2022)Li, Puig, Paxton, Du, Wang, Fan, Chen, Huang, Aky{"u}rek, Anandkumar, Andreas, Mordatch, Torralba, and Zhu}]{li2022lid}
Shuang Li, Xavier Puig, Chris Paxton, Yilun Du, Clinton Wang, Linxi Fan, Tao Chen, De-An Huang, Ekin Aky{"u}rek, Anima Anandkumar, Jacob Andreas, Igor Mordatch, Antonio Torralba, and Yuke Zhu. 2022.
\newblock Pre-trained language models for interactive decision-making.
\newblock In \emph{36th Conference on Neural Information Processing Systems (NeurIPS 2022)}.

\bibitem[{Lin et~al.(2023)Lin, Huang, Liu, Gu, Sommerer, and Ren}]{lin2023gplanet}
Bill~Yuchen Lin, Chengsong Huang, Qian Liu, Wenda Gu, Sam Sommerer, and Xiang Ren. 2023.
\newblock On grounded planning for embodied tasks with language models.
\newblock In \emph{Proceedings of the Thirty-Seventh AAAI Conference on Artificial Intelligence (AAAI-23)}.

\bibitem[{Liu et~al.(2023{\natexlab{a}})Liu, Jiang, Zhang, Liu, Zhang, Biswas, and Stone}]{liu2023llmp}
Bo~Liu, Yuqian Jiang, Xiaohan Zhang, Qiang Liu, Shiqi Zhang, Joydeep Biswas, and Peter Stone. 2023{\natexlab{a}}.
\newblock Llm+p: Empowering large language models with optimal planning proficiency.
\newblock \emph{arXiv preprint arXiv:2304.11477}.

\bibitem[{Liu et~al.(2024)Liu, Chen, Hsu, Mao, and Wu}]{liu2024learning}
Weiyu Liu, Geng Chen, Joy Hsu, Jiayuan Mao, and Jiajun Wu. 2024.
\newblock Learning planning abstractions from language.
\newblock In \emph{Proceedings of the International Conference on Learning Representations}. ICLR.
\newblock ArXiv preprint arXiv:2405.03864.

\bibitem[{Liu et~al.(2023{\natexlab{b}})Liu, Yu, Zhang, Xu, Lei, Lai, Gu, Ding, Men, Yang, Zhang, Deng, Zeng, Du, Zhang, Shen, Zhang, Su, Sun, Huang, Dong, and Tang}]{liu2023agentbenchevaluatingllmsagents}
Xiao Liu, Hao Yu, Hanchen Zhang, Yifan Xu, Xuanyu Lei, Hanyu Lai, Yu~Gu, Hangliang Ding, Kaiwen Men, Kejuan Yang, Shudan Zhang, Xiang Deng, Aohan Zeng, Zhengxiao Du, Chenhui Zhang, Sheng Shen, Tianjun Zhang, Yu~Su, Huan Sun, Minlie Huang, Yuxiao Dong, and Jie Tang. 2023{\natexlab{b}}.
\newblock \href {http://arxiv.org/abs/2308.03688} {Agentbench: Evaluating llms as agents}.

\bibitem[{Lu et~al.(2024)Lu, Liu, Liusie, Raina, Mudupalli, Zhang, and Beauchamp}]{lu2024blending}
Xiaoding Lu, Zongyi Liu, Adian Liusie, Vyas Raina, Vineet Mudupalli, Yuwen Zhang, and William Beauchamp. 2024.
\newblock \href {http://arxiv.org/abs/2401.02994} {Blending is all you need: Cheaper, better alternative to trillion-parameters llm}.

\bibitem[{Madaan et~al.(2023)Madaan, Tandon, Gupta, Hallinan, Gao, Wiegreffe, Alon, Dziri, Prabhumoye, Yang, Gupta, Majumder, Hermann, Welleck, Yazdanbakhsh, and Clark}]{madaan2023selfrefine}
Aman Madaan, Niket Tandon, Prakhar Gupta, Skyler Hallinan, Luyu Gao, Sarah Wiegreffe, Uri Alon, Nouha Dziri, Shrimai Prabhumoye, Yiming Yang, Shashank Gupta, Bodhisattwa~Prasad Majumder, Katherine Hermann, Sean Welleck, Amir Yazdanbakhsh, and Peter Clark. 2023.
\newblock \href {http://arxiv.org/abs/2303.17651} {Self-refine: Iterative refinement with self-feedback}.

\bibitem[{Moryossef et~al.(2019)Moryossef, Goldberg, and Dagan}]{moryossef-etal-2019-step}
Amit Moryossef, Yoav Goldberg, and Ido Dagan. 2019.
\newblock \href {https://doi.org/10.18653/v1/N19-1236} {{S}tep-by-step: {S}eparating planning from realization in neural data-to-text generation}.
\newblock In \emph{Proceedings of the 2019 Conference of the North {A}merican Chapter of the Association for Computational Linguistics: Human Language Technologies, Volume 1 (Long and Short Papers)}, pages 2267--2277, Minneapolis, Minnesota. Association for Computational Linguistics.

\bibitem[{O{'}Gorman et~al.(2018)O{'}Gorman, Regan, Griffitt, Hermjakob, Knight, and Palmer}]{ogorman-etal-2018-amr}
Tim O{'}Gorman, Michael Regan, Kira Griffitt, Ulf Hermjakob, Kevin Knight, and Martha Palmer. 2018.
\newblock \href {https://aclanthology.org/C18-1313} {{AMR} beyond the sentence: the multi-sentence {AMR} corpus}.
\newblock In \emph{Proceedings of the 27th International Conference on Computational Linguistics}, pages 3693--3702, Santa Fe, New Mexico, USA. Association for Computational Linguistics.

\bibitem[{Ou et~al.(2024)Ou, Uzunoglu, Durme, and Khashabi}]{ou2024worldapisworldworthapis}
Jiefu Ou, Arda Uzunoglu, Benjamin~Van Durme, and Daniel Khashabi. 2024.
\newblock \href {http://arxiv.org/abs/2407.07778} {Worldapis: The world is worth how many apis? a thought experiment}.

\bibitem[{Padmakumar et~al.(2021)Padmakumar, Thomason, Shrivastava, Lange, Narayan-Chen, Gella, Piramuthu, Tur, and Hakkani-Tur}]{padmakumar2022teach}
Aishwarya Padmakumar, Jesse Thomason, Ayush Shrivastava, Patrick Lange, Anjali Narayan-Chen, Spandana Gella, Robinson Piramuthu, Gokhan Tur, and Dilek Hakkani-Tur. 2021.
\newblock \href {http://arxiv.org/abs/2110.00534} {Teach: Task-driven embodied agents that chat}.

\bibitem[{Pallagani et~al.(2023)Pallagani, Muppasani, Murugesan, Rossi, Srivastava, Horesh, Fabiano, and Loreggia}]{pallagani2023understanding}
Vishal Pallagani, Bharath Muppasani, Keerthiram Murugesan, Francesca Rossi, Biplav Srivastava, Lior Horesh, Francesco Fabiano, and Andrea Loreggia. 2023.
\newblock Understanding the capabilities of large language models for automated planning.
\newblock \emph{arXiv preprint arXiv:2305.16151}.

\bibitem[{Pan et~al.(2024)Pan, Yaman, Nesti, Mallik, Allievi, Velipasalar, and Ren}]{pan2024vlpvisionlanguageplanning}
Chenbin Pan, Burhaneddin Yaman, Tommaso Nesti, Abhirup Mallik, Alessandro~G Allievi, Senem Velipasalar, and Liu Ren. 2024.
\newblock \href {http://arxiv.org/abs/2401.05577} {Vlp: Vision language planning for autonomous driving}.

\bibitem[{Puig et~al.(2018{\natexlab{a}})Puig, Ra, Boben, Li, Wang, Fidler, and Torralba}]{puig2018virtualhome}
Xavier Puig, Kevin Ra, Marko Boben, Jiaman Li, Tingwu Wang, Sanja Fidler, and Antonio Torralba. 2018{\natexlab{a}}.
\newblock Virtualhome: Simulating household activities via programs.
\newblock In \emph{Proceedings of the IEEE Conference on Computer Vision and Pattern Recognition}, pages 8494--8502.

\bibitem[{Puig et~al.(2018{\natexlab{b}})Puig, Ra, Boben, Li, Wang, Fidler, and Torralba}]{puig2018virtualhomesimulatinghouseholdactivities}
Xavier Puig, Kevin Ra, Marko Boben, Jiaman Li, Tingwu Wang, Sanja Fidler, and Antonio Torralba. 2018{\natexlab{b}}.
\newblock \href {http://arxiv.org/abs/1806.07011} {Virtualhome: Simulating household activities via programs}.

\bibitem[{Raman et~al.(2022)Raman, Cohen, Rosen, Idrees, Paulius, and Tellex}]{raman2022planning}
Shreyas~Sundara Raman, Vanya Cohen, Eric Rosen, Ifrah Idrees, David Paulius, and Stefanie Tellex. 2022.
\newblock Planning with large language models via corrective re-prompting.
\newblock In \emph{NeurIPS 2022 Foundation Models for Decision Making Workshop}.

\bibitem[{Russell and Norvig(1995)}]{russell95ai}
Stuart Russell and Peter Norvig. 1995.
\newblock \emph{Artificial {I}ntelligence: {A} modern approach}.
\newblock Prentice-Hall.

\bibitem[{Saparov and He(2023)}]{saparov2023languagemodelsgreedyreasoners}
Abulhair Saparov and He~He. 2023.
\newblock \href {http://arxiv.org/abs/2210.01240} {Language models are greedy reasoners: A systematic formal analysis of chain-of-thought}.

\bibitem[{Shinn et~al.(2023)Shinn, Cassano, Berman, Gopinath, Narasimhan, and Yao}]{shinn2023reflexion}
Noah Shinn, Federico Cassano, Edward Berman, Ashwin Gopinath, Karthik Narasimhan, and Shunyu Yao. 2023.
\newblock Reflexion: Language agents with verbal reinforcement learning.
\newblock In \emph{Proceedings of the 37th Conference on Neural Information Processing Systems (NeurIPS 2023)}.

\bibitem[{Shridhar et~al.(2020{\natexlab{a}})Shridhar, Thomason, Gordon, Bisk, Han, Mottaghi, Zettlemoyer, and Fox}]{shridhar2020alfredbenchmarkinterpretinggrounded}
Mohit Shridhar, Jesse Thomason, Daniel Gordon, Yonatan Bisk, Winson Han, Roozbeh Mottaghi, Luke Zettlemoyer, and Dieter Fox. 2020{\natexlab{a}}.
\newblock \href {http://arxiv.org/abs/1912.01734} {Alfred: A benchmark for interpreting grounded instructions for everyday tasks}.

\bibitem[{Shridhar et~al.(2020{\natexlab{b}})Shridhar, Yuan, C{\^o}t{\'e}, Bisk, Trischler, and Hausknecht}]{shridhar2020alfworld}
Mohit Shridhar, Xingdi Yuan, Marc-Alexandre C{\^o}t{\'e}, Yonatan Bisk, Adam Trischler, and Matthew Hausknecht. 2020{\natexlab{b}}.
\newblock Alfworld: Aligning text and embodied environments for interactive learning.
\newblock \emph{arXiv preprint arXiv:2010.03768}.

\bibitem[{Shridhar et~al.(2021)Shridhar, Yuan, Côté, Bisk, Trischler, and Hausknecht}]{shridhar2021alfworldaligningtextembodied}
Mohit Shridhar, Xingdi Yuan, Marc-Alexandre Côté, Yonatan Bisk, Adam Trischler, and Matthew Hausknecht. 2021.
\newblock \href {http://arxiv.org/abs/2010.03768} {Alfworld: Aligning text and embodied environments for interactive learning}.

\bibitem[{Silver et~al.(2023)Silver, Dan, Srinivas, Tenenbaum, Kaelbling, and Katz}]{silver2023generalized}
Tom Silver, Soham Dan, Kavitha Srinivas, Joshua~B Tenenbaum, Leslie~Pack Kaelbling, and Michael Katz. 2023.
\newblock Generalized planning in pddl domains with pretrained large language models.
\newblock \emph{arXiv preprint arXiv:2305.11014}.

\bibitem[{Silver et~al.(2022)Silver, Hariprasad, Shuttleworth, Kumar, Lozano-P{\'e}rez, and Kaelbling}]{silver2022pddl}
Tom Silver, Varun Hariprasad, Reece~S Shuttleworth, Nishanth Kumar, Tom{\'a}s Lozano-P{\'e}rez, and Leslie~Pack Kaelbling. 2022.
\newblock Pddl planning with pretrained large language models.
\newblock In \emph{NeurIPS 2022 foundation models for decision making workshop}.

\bibitem[{Singh et~al.(2022)Singh, Blukis, Mousavian, Goyal, Xu, Tremblay, Fox, Thomason, and Garg}]{singh2022progprompt}
Ishika Singh, Valts Blukis, Arsalan Mousavian, Ankit Goyal, Danfei Xu, Jonathan Tremblay, Dieter Fox, Jesse Thomason, and Animesh Garg. 2022.
\newblock Progprompt: Generating situated robot task plans using large language models.
\newblock In \emph{Proceedings of the 36th Conference on Neural Information Processing Systems (NeurIPS)}.

\bibitem[{Song et~al.(2023)Song, Wu, Washington, Sadler, Chao, and Su}]{song2023llmplanner}
Chan~Hee Song, Jiaman Wu, Clayton Washington, Brian~M. Sadler, Wei-Lun Chao, and Yu~Su. 2023.
\newblock Llm-planner: Few-shot grounded planning for embodied agents with large language models.
\newblock In \emph{Proceedings of the IEEE International Conference on Computer Vision (ICCV)}.

\bibitem[{Stent et~al.(2004)Stent, Prasad, and Walker}]{stent-etal-2004-trainable}
Amanda Stent, Rashmi Prasad, and Marilyn Walker. 2004.
\newblock \href {https://doi.org/10.3115/1218955.1218966} {Trainable sentence planning for complex information presentations in spoken dialog systems}.
\newblock In \emph{Proceedings of the 42nd Annual Meeting of the Association for Computational Linguistics ({ACL}-04)}, pages 79--86, Barcelona, Spain.

\bibitem[{Sun et~al.(2023)Sun, Zhuang, Kong, Dai, and Zhang}]{sun2023adaplanner}
Haotian Sun, Yuchen Zhuang, Lingkai Kong, Bo~Dai, and Chao Zhang. 2023.
\newblock \href {http://arxiv.org/abs/2305.16653} {Adaplanner: Adaptive planning from feedback with language models}.

\bibitem[{Valmeekam et~al.(2023)Valmeekam, Marquez, Olmo, Sreedharan, and Kambhampati}]{valmeekam2023planbench}
Karthik Valmeekam, Matthew Marquez, Alberto Olmo, Sarath Sreedharan, and Subbarao Kambhampati. 2023.
\newblock Planbench: An extensible benchmark for evaluating large language models on planning and reasoning about change.
\newblock In \emph{Thirty-seventh Conference on Neural Information Processing Systems Datasets and Benchmarks Track}.

\bibitem[{Wang et~al.(2023{\natexlab{a}})Wang, Todd, Yuan, Xiao, Côté, and Jansen}]{wang2023bytesized32corpuschallengetask}
Ruoyao Wang, Graham Todd, Eric Yuan, Ziang Xiao, Marc-Alexandre Côté, and Peter Jansen. 2023{\natexlab{a}}.
\newblock \href {http://arxiv.org/abs/2305.14879} {Bytesized32: A corpus and challenge task for generating task-specific world models expressed as text games}.

\bibitem[{Wang et~al.(2023{\natexlab{b}})Wang, Cai, Chen, Liu, Ma, and Liang}]{wang2023describe}
Zihao Wang, Shaofei Cai, Guanzhou Chen, Anji Liu, Xiaojian Ma, and Yitao Liang. 2023{\natexlab{b}}.
\newblock \href {https://github.com/CraftJarvis/MC-Planner} {Describe, explain, plan and select: Interactive planning with large language models}.
\newblock In \emph{37th Conference on Neural Information Processing Systems (NeurIPS)}.

\bibitem[{Wong et~al.(2023)Wong, Mao, Sharma, Siegel, Feng, Korneev, Tenenbaum, and Andreas}]{wong2023learning}
Lionel Wong, Jiayuan Mao, Pratyusha Sharma, Zachary~S. Siegel, Jiahai Feng, Noa Korneev, Joshua~B. Tenenbaum, and Jacob Andreas. 2023.
\newblock \href {http://arxiv.org/abs/2312.08566} {Learning adaptive planning representations with natural language guidance}.

\bibitem[{Wu et~al.(2024)Wu, Tang, Mitchell, and Li}]{wu2024smartplaybenchmarkllmsintelligent}
Yue Wu, Xuan Tang, Tom~M. Mitchell, and Yuanzhi Li. 2024.
\newblock \href {http://arxiv.org/abs/2310.01557} {Smartplay: A benchmark for llms as intelligent agents}.

\bibitem[{Xie et~al.(2024)Xie, Zhang, Chen, Zhu, Lou, Tian, Xiao, and Su}]{xie2024travelplannerbenchmarkrealworldplanning}
Jian Xie, Kai Zhang, Jiangjie Chen, Tinghui Zhu, Renze Lou, Yuandong Tian, Yanghua Xiao, and Yu~Su. 2024.
\newblock \href {http://arxiv.org/abs/2402.01622} {Travelplanner: A benchmark for real-world planning with language agents}.

\bibitem[{Yang et~al.(2024)Yang, Chen, Zhang, Yuan, Chen, Richardson, Xiao, and Yang}]{yang2024selfgoallanguageagentsknow}
Ruihan Yang, Jiangjie Chen, Yikai Zhang, Siyu Yuan, Aili Chen, Kyle Richardson, Yanghua Xiao, and Deqing Yang. 2024.
\newblock \href {http://arxiv.org/abs/2406.04784} {Selfgoal: Your language agents already know how to achieve high-level goals}.

\bibitem[{Yao et~al.(2023{\natexlab{a}})Yao, Yu, Zhao, Shafran, Griffiths, Cao, and Narasimhan}]{yao2023treethoughtsdeliberateproblem}
Shunyu Yao, Dian Yu, Jeffrey Zhao, Izhak Shafran, Thomas~L. Griffiths, Yuan Cao, and Karthik Narasimhan. 2023{\natexlab{a}}.
\newblock \href {http://arxiv.org/abs/2305.10601} {Tree of thoughts: Deliberate problem solving with large language models}.

\bibitem[{Yao et~al.(2023{\natexlab{b}})Yao, Zhao, Yu, Du, Shafran, Narasimhan, and Cao}]{yao2023react}
Shunyu Yao, Jeffrey Zhao, Dian Yu, Nan Du, Izhak Shafran, Karthik Narasimhan, and Yuan Cao. 2023{\natexlab{b}}.
\newblock React: Synergizing reasoning and acting in language models.
\newblock In \emph{Proceedings of the International Conference on Learning Representations (ICLR 2023)}.

\bibitem[{Yuan et~al.(2023{\natexlab{a}})Yuan, Kazemi, Xu, Noble, Imbrasaite, and Ramachandran}]{yuan2023tasklamaprobingcomplextask}
Quan Yuan, Mehran Kazemi, Xin Xu, Isaac Noble, Vaiva Imbrasaite, and Deepak Ramachandran. 2023{\natexlab{a}}.
\newblock \href {http://arxiv.org/abs/2308.15299} {Tasklama: Probing the complex task understanding of language models}.

\bibitem[{Yuan et~al.(2023{\natexlab{b}})Yuan, Chen, Fu, Ge, Shah, Jankowski, Xiao, and Yang}]{yuan2023distillingscriptknowledgelarge}
Siyu Yuan, Jiangjie Chen, Ziquan Fu, Xuyang Ge, Soham Shah, Charles~Robert Jankowski, Yanghua Xiao, and Deqing Yang. 2023{\natexlab{b}}.
\newblock \href {http://arxiv.org/abs/2305.05252} {Distilling script knowledge from large language models for constrained language planning}.

\bibitem[{Zhang et~al.(2024)Zhang, Yin, Wang, and Xiang}]{zhang2024itcma}
Hanzhong Zhang, Jibin Yin, Haoyang Wang, and Ziwei Xiang. 2024.
\newblock Itcma: A generative agent based on a computational consciousness structure.
\newblock \emph{arXiv preprint arXiv:2403.20097}.

\bibitem[{Zhang et~al.(2021)Zhang, Jauhar, Kiseleva, White, and Roth}]{zhang2021learning}
Yi~Zhang, Sujay~Kumar Jauhar, Julia Kiseleva, Ryen White, and Dan Roth. 2021.
\newblock Learning to decompose and organize complex tasks.
\newblock In \emph{Proceedings of the 2021 Conference of the North American Chapter of the Association for Computational Linguistics: Human Language Technologies}, pages 2726--2735.

\bibitem[{Zhao et~al.(2023)Zhao, Wang, Yu, and Mei}]{zhao2023explicit}
Hongyu Zhao, Kangrui Wang, Mo~Yu, and Hongyuan Mei. 2023.
\newblock Explicit planning helps language models in logical reasoning.
\newblock \emph{arXiv preprint arXiv:2303.15714}.

\bibitem[{Zheng et~al.(2024)Zheng, Mishra, Zhang, Chen, Chen, Nova, Hou, Cheng, Le, Chi, and Zhou}]{zheng2024naturalplanbenchmarkingllms}
Huaixiu~Steven Zheng, Swaroop Mishra, Hugh Zhang, Xinyun Chen, Minmin Chen, Azade Nova, Le~Hou, Heng-Tze Cheng, Quoc~V. Le, Ed~H. Chi, and Denny Zhou. 2024.
\newblock \href {http://arxiv.org/abs/2406.04520} {Natural plan: Benchmarking llms on natural language planning}.

\bibitem[{Zhou et~al.(2023)Zhou, Song, Yao, Shu, and Ma}]{zhou2023isrllm}
Zhehua Zhou, Jiayang Song, Kunpeng Yao, Zhan Shu, and Lei Ma. 2023.
\newblock Isr-llm: Iterative self-refined large language model for long-horizon sequential task planning.
\newblock \emph{arXiv preprint arXiv:2308.13724}.

\end{thebibliography}
\bibliographystyle{acl_natbib}








  

\end{document}